\documentclass[runningheads]{llncs}

\usepackage{hyperref}

\usepackage{color}

\usepackage[T1]{fontenc}
\usepackage{booktabs}
\usepackage{amsmath}
\usepackage{multirow}
\usepackage{enumitem}
\usepackage{amsmath}
\usepackage{amsfonts}
\usepackage{placeins}
\usepackage{float}

\usepackage{enumitem}

\newcommand{\sstitle}[1]{\smallskip\noindent\textbf{#1.\/}}

\usepackage{epstopdf}
\usepackage{graphics}
\DeclareGraphicsExtensions{.eps,.gz,.eps,.pdf,.png,.jpg}
\graphicspath{{./}{./figures/}{../figures/}}

\usepackage{graphicx}

\begin{document}

\title{Empowering Contrastive Federated Sequential Recommendation with LLMs}

\titlerunning{LUMOS: LLM-Driven Contrastive FedSeqRec}

\author{
Thi Minh Chau Nguyen
\and
Minh Hieu Nguyen
\and
Duc Anh Nguyen
\and
Xuan Huong Tran
\and
Thanh Trung Huynh
\and
Quoc Viet Hung Nguyen
}

\authorrunning{T.M.C. Nguyen et al.}


\maketitle

\begin{abstract}
Federated sequential recommendation (FedSeqRec) aims to perform next-item prediction while keeping user data decentralised, yet model quality is frequently constrained by fragmented, noisy, and homogeneous interaction logs stored on individual devices. Many existing approaches attempt to compensate through manual data augmentation or additional server-side constraints, but these strategies either introduce limited semantic diversity or increase system overhead. To overcome these challenges, we propose \textbf{LUMOS}, a parameter-isolated FedSeqRec architecture that integrates large language models (LLMs) as \emph{local semantic generators}. Instead of sharing gradients or auxiliary parameters, LUMOS privately invokes an on-device LLM to construct three complementary sequence variants from each user history: (i) \emph{future-oriented} trajectories that infer plausible behavioural continuations, (ii) \emph{semantically equivalent rephrasings} that retain user intent while diversifying interaction patterns, and (iii) \emph{preference-inconsistent counterfactuals} that serve as informative negatives. These synthesized sequences are jointly encoded within the federated backbone through a tri-view contrastive optimisation scheme, enabling richer representation learning without exposing sensitive information. Experimental results across three public benchmarks show that LUMOS achieves consistent gains over competitive centralised and federated baselines on HR@20 and NDCG@20. In addition, the use of semantically grounded positive signals and counterfactual negatives improves robustness under noisy and adversarial environments, even without dedicated server-side protection modules. Overall, this work demonstrates the potential of LLM-driven semantic generation as a new paradigm for advancing privacy-preserving federated recommendation.

\end{abstract}

\setlength{\belowdisplayskip}{2pt}
\setlength{\belowdisplayshortskip}{2pt}
\setlength{\abovedisplayskip}{2pt}
\setlength{\abovedisplayshortskip}{2pt}

\section{Introduction}

Personalized services such as e-commerce, news feeds, and online media platforms increasingly rely on models that can understand how user interests evolve over time~\cite{nguyen2025ondevice}. Sequential recommender systems (SeqRecs) are designed for this purpose: instead of treating interactions as unordered histories, they explicitly model ordered user--item sequences to infer the next item a user is likely to engage with~\cite{rw_15,rw_17}. Early work extended traditional collaborative filtering~\cite{rw_11,rw_12,rw_13} by incorporating simple temporal dependencies, while later approaches adopted neural architectures to capture complex sequential patterns. Representative examples include Markov-chain-based models~\cite{rw_14}, recurrent architectures such as GRU4Rec~\cite{rw_16}, and self-attention-based methods like SASRec~\cite{rw_17} and Cloze-style sequence models~\cite{rw_18}. More recently, ID-free sequential recommenders have exploited large pre-trained language models (PLMs) such as BERT~\cite{rw_112} and larger LLMs~\cite{rw_110,rw_111,rw_113,nguyen2026review} to encode interaction sequences purely from textual signals.

Most of these sequential models, however, assume that all user interaction logs are collected and stored centrally. Federated learning (FL) offers a natural alternative by training models collaboratively across distributed clients without exposing raw data~\cite{rw_22}. When FL is combined with sequential recommendation, we obtain Federated Sequential Recommendation (FedSeqRec), where each client maintains its own interaction history and participates in global model updates by contributing local gradients or predictions~\cite{fmss,ptf-fsr}. We refer to concrete systems that operate under this paradigm as Federated Sequential Recommendation Systems (FedSRS).

Despite promising progress, current FedSRS designs remain limited in how effectively they exploit users' local sequences. First, the amount of data available on each client is often very small, highly sparse, and noisy, especially for cold or infrequent users. Training a powerful sequential encoder in such conditions can lead to unstable representations and poor generalization. Second, although contrastive learning has proved to be an effective self-supervised signal in centralized recommendation~\cite{rw_41,rw_42,rw_49}, directly transplanting these techniques into FedSeqRec is non-trivial. Typical data-based augmentations (e.g., random cropping, shuffling, item masking) or representation-based noise injection were designed under the assumption of large centralized corpora~\cite{rw_410,rw_411,rw_412,rw_44,rw_45,rw_46,rw_415}. When applied to short, client-specific sequences, such heuristics may either destroy the semantic structure of user behavior or yield weak, nearly redundant views. Third, preserving strong privacy guarantees in FL restricts access to cross-client information, making it difficult to construct diverse contrastive pairs without leaking sensitive content or transmitting model parameters~\cite{rw_51,rw_53,rw_54}.

At the same time, large language models have shown strong ability to interpret 
and generate coherent behavioral trajectories when interactions are expressed in 
natural language~\cite{rw_110,rw_111,rw_113}. As a behavioral prior, an LLM can 
infer plausible futures, paraphrase equivalent preference patterns, or propose 
realistic counterfactuals from a partial sequence. This leads to a central 
question: \emph{can we exploit client-side LLMs to synthesize richer training 
signals for FedSeqRec without altering the FL protocol or weakening privacy 
guarantees?}

In this paper, we answer this question by introducing LUMOS, a federated sequential recommendation framework that integrates LLM-driven behavioral synthesis with multi-view contrastive learning. In LUMOS, each client treats an LLM as a local behavioral generator. Conditioned on the user's observed interaction history, the LLM produces auxiliary sequences that (i) maintain the user's intent while varying surface-level patterns, or (ii) represent hard yet realistic negatives that conflict with the user's true preferences. These synthetic sequences are never uploaded to the server; instead, they serve as additional views for a contrastive objective that is optimized entirely on-device. The global server only aggregates model updates following the standard FL protocol. Unlike prior FedSRS work that explores robustness against adversarial attacks~\cite{rw_31,rw_32,a-hum,psmu}, our focus here is on enhancing representation quality and data efficiency in benign federated environments, leveraging LLMs as semantic augmenters rather than as attack or defense mechanisms.
We evaluate LUMOS on three real-world benchmarks. Experimental results show that LUMOS consistently improves top-$K$ accuracy over both centralized and federated baselines, especially in regimes with sparse and short client histories. Ablation studies further demonstrate the contribution of LLM-generated views and the multi-view contrastive objective.

In summary, this work makes the following contributions:
\begin{itemize}[noitemsep]
    \item We present LUMOS, a federated sequential recommendation framework that harnesses LLMs as on-device behavioral generators to enrich self-supervised signals in FedSeqRec.
    \item We propose a privacy-preserving, multi-view contrastive learning scheme that jointly leverages real and LLM-synthesized sequences on each client, without transmitting model parameters, embeddings, or synthetic data.
    \item We instantiate LUMOS with several widely used sequential backbones and conduct extensive experiments on three public datasets, showing consistent gains over state-of-the-art federated baselines.
\end{itemize}

\section{Related Works}
\label{sec:related}

\sstitle{Sequential Recommender Systems}
User behaviors in modern online platforms (e.g., shopping, reading, media consumption) often form temporal sequences that reveal evolving user interests. Early collaborative-filtering recommenders~\cite{rw_11,rw_12,rw_13} treated interactions as unordered sets, limiting their ability to model temporal dynamics. To address this, Rendle et al.~\cite{rw_14} introduced one of the first sequential models based on Markov chains. With the rise of deep neural networks, sequential recommendation quickly adopted recurrent architectures~\cite{rw_15}, such as GRU4Rec~\cite{rw_16}, and later attention-based structures like SASRec~\cite{rw_17}, which demonstrated strong capabilities in capturing long-range dependencies. Sun et al.~\cite{rw_18} extended this line of work by incorporating a Cloze-style objective, enabling bidirectional context modeling. The emergence of large language models (LLMs) has sparked a new direction in sequential recommendation. Instead of relying solely on item IDs, ID-free approaches leverage textual semantics and natural-language modeling to represent user histories. Subsequent LLM-driven models~\cite{rw_113} showed that large pre-trained language models, including BERT~\cite{rw_112} and more powerful LLM variants, can effectively encode preference sequences without specialized ID embeddings. This work builds on these insights and considers LLMs as auxiliary behavioral generators that can enhance sequence modeling in federated environments.

\sstitle{Federated Recommender Systems}
Federated recommendation has emerged to reconcile personalization with data protection requirements~\cite{rw_21,rw_22,rw_23}. The earliest frameworks, such as the federated collaborative filtering model by Ammad et al.~\cite{rw_24}, demonstrated the feasibility of joint recommendation without sharing raw interactions. Subsequent advancements explored better optimization strategies~\cite{rw_25,rw_26,rw_27}, improved privacy preservation~\cite{long2024physical,qu2024towards,rw_210,rw_211,rw_212}, robustness considerations~\cite{rw_31,rw_32}, and data management~\cite{nguyen2023poisoning,nguyen2023example,nguyen2014reconciling,nguyen2015smart,thang2015evaluation,nguyen2015tag,hung2019handling}.

\sstitle{LLMs for Recommendation and Federated LLMs}
Large language models have recently been incorporated into recommender systems in multiple ways. Lyu et al.~\cite{llmrec} prompt an LLM to rewrite item descriptions, improving centralized recommendation quality. Wu et al.~\cite{wu2024llm4rec_survey} and Shehmir and Kashef~\cite{fi17060252} survey this LLM4Rec landscape, covering discriminative vs.\ generative uses, adaptation strategies (fine-tuning vs.\ prompting), and challenges such as scalability, latency, and privacy. Tang et al.~\cite{tang2025} further suggest that a single LLM can act as a domain-agnostic recommender.
In parallel, federated LLM learning (FedLLM) has emerged as a distinct area: Yao et al.~\cite{yao2024} review techniques for training LLMs under heterogeneous constraints, and Ye et al.~\cite{ye2024} introduce FedLLM-Bench for benchmarking on realistic user-split datasets.
Our work relates to both LLM4Rec and FedLLM but differs in how the LLM is used. Instead of federating the LLM~\cite{yao2024,ye2024} or applying it only as a text encoder or centralized recommender~\cite{llmrec,wu2024llm4rec_survey,fi17060252,tang2025}, we assume a pre-trained LLM is available locally on each client and treat it as a semantic behavioral generator. Conditioned on a user's history, the LLM produces auxiliary, behaviorally plausible interaction sequences that we use as additional contrastive views for federated sequential recommendation. To our knowledge, this is the first work to leverage LLM-generated interaction sequences as on-device contrastive views in FedSeqRec~\cite{zhao2021eires,huynh2021network,duong2022deep,nguyen2022model,nguyen2023detecting,trung2022learning,huynh2023efficient}.

\sstitle{Contrastive Learning in Federated Recommendation}
Contrastive learning in FedRecs remains relatively underexplored. FedCL~\cite{rw_51} incorporated contrastive objectives by sending model parameters to the server, relying on local differential privacy~\cite{rw_52} to mitigate leakage. Later approaches~\cite{rw_53,rw_54} follow a similar principle of exchanging embeddings or statistics, which weakens the privacy guarantees of federated training. Meanwhile, UNION~\cite{rw_37} applies contrastive ideas to detect untargeted attacks rather than improve representation quality under benign settings. Our work differs by preserving strict FL constraints. Instead, clients rely on LLM-generated behavioral views to enrich training via multi-view contrastive learning. This preserves privacy while substantially improving sequential representations in data-scarce federated environments~\cite{thang2022nature,duong2022efficient,nguyen2020factcatch,hung2017answer,nguyen2017argument,ren2022prototype,nguyen2018if,toan2018diversifying}.

\section{Model and Problem Formulation}
\label{sec:preliminaries}

\sstitle{Problem Setting}
We study \emph{federated sequential recommendation}, where a set of users $U = \{u_1, \dots, u_N\}$ interact with a global item corpus $V = \{v_1, \dots, v_M\}$.  
Each user $u \in U$ holds a private, time-ordered interaction sequence
\[
S_u = [v_{u,1}, v_{u,2}, \dots, v_{u,T_u}],
\]
which remains entirely on the client device. 
The goal is to learn a sequential recommendation model $f_\theta(\cdot)$ that 
predicts the next-item preference distribution:
\[
\hat{y}_u = f_\theta(S_u),
\]
from which the next item $v_{u,T_u+1}$ is ranked among the candidates in $V$.

\sstitle{Federated Optimization Protocol}
We follow a standard synchronous federated learning protocol. 
In each round $t$, the server holds a global model $\theta^{(t)}$ and broadcasts 
it to a sampled user subset $\mathcal{U}_t \subseteq U$. Each client 
$u \in \mathcal{U}_t$ initializes its local model with $\theta^{(t)}$ and runs 
$E$ epochs of stochastic gradient descent on its private data $S_u$, producing 
updated parameters $\theta_u^{(t)}$. The server then aggregates these updates 
using the FedAvg rule:
\[
\theta^{(t+1)} = \frac{1}{|\mathcal{U}_t|} \sum_{u \in \mathcal{U}_t} \theta^{(t)}_u.
\]
Throughout this process, raw interaction sequences, LLM prompts, and LLM-generated texts stay on-device; only model parameters are exchanged.

\sstitle{Sequential Recommendation Backbone}
Following prior work on sequential recommendation~\cite{rw_16,rw_17}, we assume $f_\theta(\cdot)$ is instantiated as a neural sequence encoder such as GRU4Rec~\cite{rw_16} or SASRec~\cite{rw_17}. Given a sequence $S_u$, the encoder produces a user representation $h_u = f_\theta(S_u)$, which is then projected to a score vector over items. The standard next-item prediction loss for user $u$ is:
\[
\mathcal{L}^{\mathrm{Rec}}_u
= - \log p_\theta\bigl(v_{u,T_u+1} \mid S_u\bigr),
\]
where $p_\theta(\cdot)$ denotes the predicted probability distribution over $V$.

\sstitle{Client-Side Data Scarcity}
In federated settings, each client typically observes only a short, sparse interaction history, especially for cold-start or infrequent users. This makes it difficult to construct informative self-supervised signals using conventional augmentations such as masking, cropping, or shuffling, which assume longer sequences and richer behavioral contexts. As a result, purely local training can lead to unstable user embeddings.

\sstitle{LLM-Generated Behavioral Views}
To enrich local supervision without sharing data, each client accesses a 
pre-trained large language model (LLM) as a black-box generator. Conditioned 
on the original sequence $S_u$, the client privately queries the LLM to obtain 
several synthetic behavioral variants:
\[
S_u^{\mathrm{F}}, \quad
S_u^{\mathrm{P}}, \quad
S_u^{\mathrm{N}},
\]
which, at a high level, correspond to \emph{predictive future}, \emph{intent-preserving paraphrased}, and \emph{counterfactual negative} views of the user’s behavior. These sequences are never transmitted to the server; they are used solely to construct richer training signals on the client side. The same backbone $f_\theta(\cdot)$ encodes these sequences into representations:
\[
h_u^{\mathrm{F}} = f_\theta(S_u^{\mathrm{F}}), \quad
h_u^{\mathrm{P}} = f_\theta(S_u^{\mathrm{P}}), \quad
h_u^{\mathrm{N}} = f_\theta(S_u^{\mathrm{N}}).
\]

\sstitle{Client Objective and Global Goal}
During local training, each participating client $u$ optimizes a composite objective that combines supervised prediction with tri-view contrastive learning:
\[
\mathcal{L}_u
=
\mathcal{L}^{\mathrm{Rec}}_u
+
\lambda_{\mathrm{CL}} \, \mathcal{L}^{\mathrm{CL}}_u,
\]
where $\mathcal{L}^{\mathrm{CL}}_u$ is the tri-view contrastive loss defined in \autoref{sec:methodology}, and $\lambda_{\mathrm{CL}}$ controls its strength relative to the supervised term.

The federated training objective is then:
\[
\min_{\theta} \;
\sum_{u \in U} \mathbb{E}\bigl[\mathcal{L}_u(\theta)\bigr],
\]
where the expectation is taken over the stochastic client sampling $\mathcal{U}_t$ and the randomness of LLM generation and optimization. Under this formulation, the server optimizes $\theta$ purely through aggregated model updates, while all raw and synthetic behavioral data remain strictly local.

\section{Methodology}
\label{sec:methodology}

\begin{figure*}[t]  %
    \centering
    \includegraphics[width=0.8\linewidth]{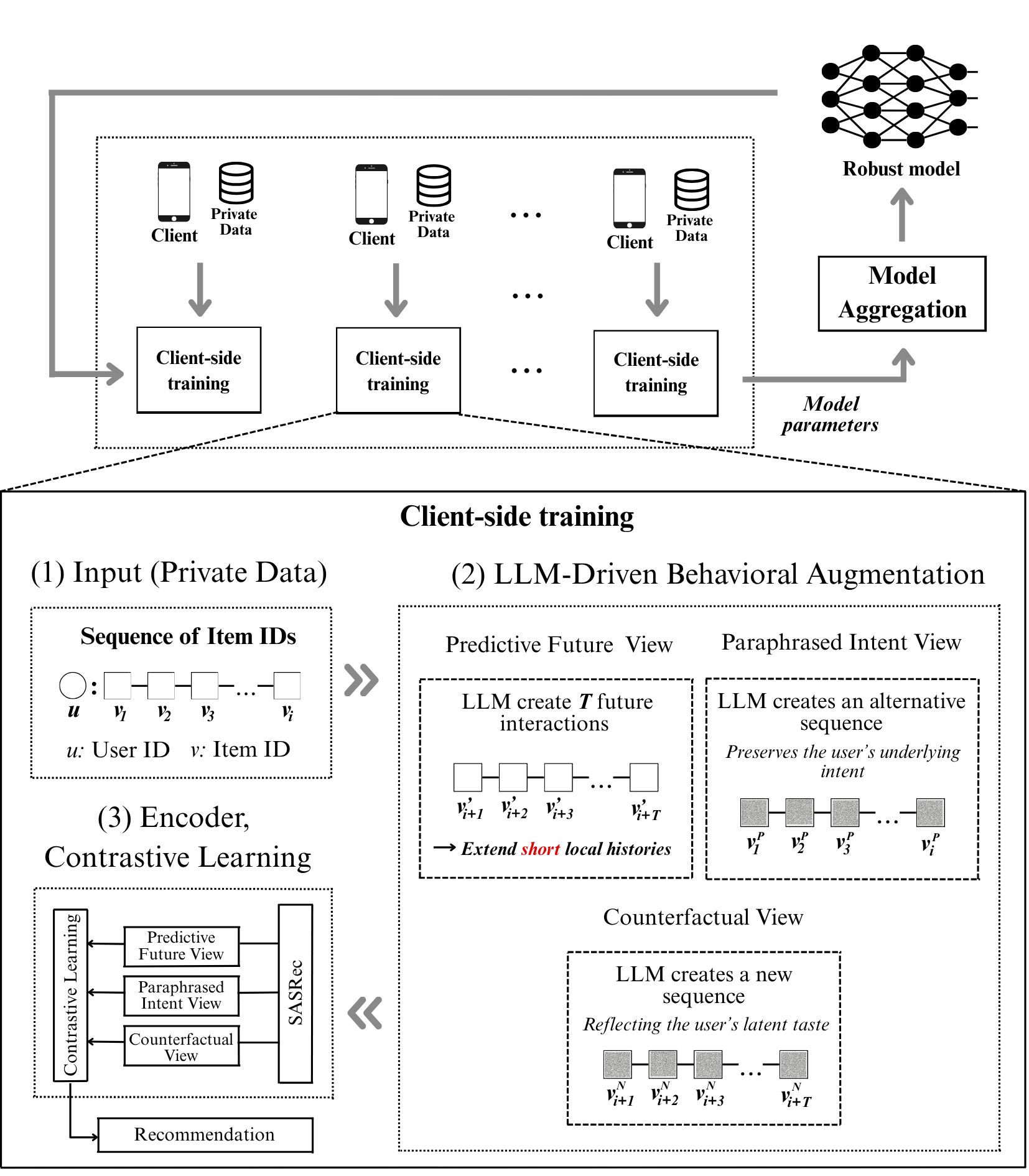}
    \caption{Overall architecture of the proposed LUMOS framework.}
    \label{fig:framework}
\end{figure*}

In this section, we present LUMOS, a federated sequential recommendation framework that enriches local training with LLM-generated behavioral views. The overall architecture of LUMOS is illustrated in \autoref{fig:framework}. The key idea is to compensate for the severe data sparsity on each client by using a pre-trained LLM as a local behavioral generator, and then integrating these synthetic sequences into a tri-view contrastive learning objective. The FL protocol itself remains standard FedAvg; all novelty lies in how each client constructs and exploits additional semantic views during local optimization.

\subsection{Federated Sequential Recommendation Setup}

We adopt the federated setting described in \autoref{sec:preliminaries}. Each user $u$ owns a private sequence $S_u = [v_{u,1}, \dots, v_{u,T_u}]$ and the shared backbone $f_\theta(\cdot)$ (e.g., GRU4Rec~\cite{rw_16} or SASRec~\cite{rw_17}) encodes $S_u$ into a representation $h_u = f_\theta(S_u)$. The next-item prediction loss is:
\[
\mathcal{L}^{\mathrm{Rec}}_u
= - \log p_\theta\bigl(v_{u,T_u+1} \mid S_u\bigr),
\]
and model updates are aggregated on the server via FedAvg~\cite{rw_22}. Raw and synthetic sequences never leave the device.

\subsection{LLM-Driven Behavioral Augmentation}
\label{sec:method_llm_views}

In standard FedSeqRec, each client typically has only a few interactions, which 
limits the effectiveness of conventional augmentations such as masking or random 
reordering. To mitigate this, LUMOS assumes that each client can query a 
pre-trained LLM locally (or via a privacy-preserving API) and use it to generate 
additional behavioral views conditioned on $S_u$. We consider three such views.

\sstitle{Predictive Future View} 
This view extrapolates plausible future behaviors beyond the user's last recorded interaction. Given the sequence $S_u$, the client queries the LLM to produce a short continuation that represents likely next items for the same user, generating an extended future-oriented sequence.
\[
S_u^{\mathrm{F}} = [v_{u,T_u+1}', v_{u,T_u+2}', \dots].
\]

\sstitle{Paraphrased Intent-Preserving View} 
This view captures alternative yet semantically equivalent ways of expressing the same underlying preferences. 
Given the original sequence $S_u$, the client asks the LLM to rewrite it while preserving the user's intent -- for example, by substituting items with close alternatives or applying mild reordering -- to produce a coherent paraphrased sequence.
\[
S_u^{\mathrm{P}} = [\tilde{v}_{u,1}, \tilde{v}_{u,2}, \dots].
\]

\sstitle{Counterfactual Negative View}
The counterfactual view synthesizes behaviors that are intentionally misaligned with the user’s preferences. Conditioned on $S_u$, the client asks the LLM to generate a sequence that is unlikely for this user (e.g., from opposite categories or unrelated domains), giving:
\[
S_u^{\mathrm{N}} = [v_{u,1}^{-}, v_{u,2}^{-}, \dots].
\]

\sstitle{Encoding All Views}
All sequences are encoded by the same backbone:
\[
h_u = f_\theta(S_u), \quad
h_u^{\mathrm{F}} = f_\theta(S_u^{\mathrm{F}}), \quad
h_u^{\mathrm{P}} = f_\theta(S_u^{\mathrm{P}}), \quad
h_u^{\mathrm{N}} = f_\theta(S_u^{\mathrm{N}}).
\]
These representations form the basis of our tri-view contrastive objective.

\subsection{Tri-View Contrastive Learning}
\label{sec:method_cl}

Contrastive learning has proved effective for recommendation in centralized settings~\cite{rw_41,rw_42,rw_49}. In LUMOS, we adapt this idea to the federated sequential scenario by constructing a tri-view objective that leverages the LLM-generated views.

For each user $u$, we treat the real sequence $h_u$ as the anchor and define two positive views (future and paraphrase) and one negative view (counterfactual):
\[
\mathcal{P}_u = \{(h_u, h_u^{\mathrm{F}}), \, (h_u, h_u^{\mathrm{P}})\}, 
\qquad
\mathcal{N}_u = \{(h_u, h_u^{\mathrm{N}})\}.
\]
We define the positive and negative partition functions as:
\begin{align}
\mathrm{pos}(u) &= \sum_{(h_a,h_p)\in\mathcal{P}_u}
\exp(\mathrm{sim}(h_a,h_p)/\tau), \\
\mathrm{neg}(u) &= \sum_{(h_a,h_n)\in\mathcal{N}_u}
\exp(\mathrm{sim}(h_a,h_n)/\tau).
\end{align}
The tri-view contrastive loss is then:
\begin{equation}
\label{eq:tri_view_cl}
\mathcal{L}^{\mathrm{CL}}_u
=
-\log
\frac{\mathrm{pos}(u)}{\mathrm{pos}(u) + \mathrm{neg}(u)}.
\end{equation}
Here $\mathrm{sim}(\cdot,\cdot)$ is a similarity function (e.g., cosine similarity) and $\tau$ is a temperature hyperparameter. Intuitively, this loss encourages $h_u$ to be close to the future and paraphrased views while being far from the counterfactual view. Because all views are generated conditionally on $S_u$ by the LLM, the resulting positives and negatives are semantically rich and user-specific, which is particularly valuable when $S_u$ is short.

\subsection{Local Objective and Federated Training}

On each participating client, the overall training objective combines supervised prediction with tri-view contrastive learning:
\[
\mathcal{L}_u
=
\mathcal{L}^{\mathrm{Rec}}_u
+
\lambda_{\mathrm{CL}} \, \mathcal{L}^{\mathrm{CL}}_u,
\]
where $\lambda_{\mathrm{CL}}$ controls the strength of the contrastive term. Each client initializes its local model with the global parameters $\theta^{(t)}$, runs several epochs of SGD to minimize $\mathcal{L}_u$, and uploads the updated parameters $\theta^{(t)}_u$ to the server. The server then computes the new global model by averaging:
\[
\theta^{(t+1)} = \frac{1}{|\mathcal{U}_t|} \sum_{u \in \mathcal{U}_t} \theta^{(t)}_u.
\]

Under LUMOS, all LLM calls, synthetic sequences, and tri-view representations are confined to the client side. The server remains agnostic to the details of augmentation and simply orchestrates standard federated optimization, while the client-side objectives leverage LLM-generated semantic views to learn richer sequential representations from sparse interaction histories.
\section{Experiments}
\label{sec:experiments}

We empirically evaluate LUMOS on three real-world benchmarks and address the following questions:
\begin{enumerate}[label=\textbf{RQ\arabic*:}, leftmargin=40pt, labelsep=0.5em, itemsep=0pt]
    \item How does LUMOS compare with state-of-the-art recommenders?
    \item How much do LLM-driven views contribute to performance?
    \item How sensitive is LUMOS to key hyperparameters?
\end{enumerate}

\subsection{Experimental Setup}

\sstitle{Datasets}
We use three standard sequential recommendation datasets: 
Amazon Cell Phone, Amazon Baby, and MIND~\cite{dataset_amazon,dataset_mind}, 
covering electronics, baby products, and news. 
Following common practice~\cite{rw_16,rw_17}, we remove users and items 
with fewer than five interactions, sort each user's interactions by timestamp, 
reserve the last two interactions for validation and testing, and cap the 
maximum sequence length at $50$ (discarding sequences shorter than $5$). 
Each remaining user corresponds to a single client in the federated setting.

\sstitle{Evaluation protocol}
We adopt Hit Ratio at rank 20 (HR@20) and Normalized Discounted Cumulative Gain at rank 20 (NDCG@20) \cite{rw_16,rw_17}. HR@20 measures whether the ground-truth next item appears in the top-20 list, while NDCG@20 also accounts for the item’s position. For each test user, we rank all candidate items not previously interacted with and compute metrics over the full ranking to avoid sampling bias~\cite{krichene2020sampled}.

\sstitle{Baselines}
We compare LUMOS against both centralized and FedSeqRec:
\begin{itemize}
    \item \textbf{Centralized SRS.} GRU4Rec~\cite{rw_16} is the classical GRU-based session model, and SASRec~\cite{rw_17} is an unidirectional self-attention sequential recommender
    \item \textbf{Standard federated SeqRec (FedSRS).}
FedSeq(GRU4Rec) and FedSeq(SASRec) are vanilla FedAvg versions of GRU4Rec and SASRec, respectively: each client trains its backbone locally and the server aggregates parameters, without any contrastive learning or LLM augmentation.
\item \textbf{Contrastive federated SRS (ConFedSRS).}
ConFedSRS(GRU4Rec) and ConFedSRS(SASRec) extend FedSRS by adding a standard sequence-level contrastive loss on locally augmented sequences (e.g., masking and cropping), without using LLMs.
\item \textbf{LUMOS (ours).}
LUMOS is a federated model with a SASRec backbone that never leaves the client. Each client uses a local (or privacy-preserving) LLM to generate three behavioral views (future, paraphrased, counterfactual) for its own sequence and optimizes a tri-view contrastive loss in addition to next-item prediction.
\end{itemize}

\sstitle{Implementation details}
All models are implemented in PyTorch and trained on a server with NVIDIA A100 GPUs. Unless otherwise specified, LUMOS uses SASRec as its sequential backbone. We adopt synchronous FedAvg, sampling 10\% of clients per communication round; each selected client runs five local epochs before aggregation, and we train for 100 rounds. We use Adam with learning rate $1\times 10^{-3}$, weight decay $1\times 10^{-5}$, batch size 128, and gradient clipping with maximum norm 5.0. The contrastive loss weight is set to $\lambda_{\text{CL}} = 0.1$ and the temperature to $\tau = 0.07$.

\subsection{Overall Recommendation Performance (RQ1)}

Table~\ref{tab:overall-performance} reports HR@20 and NDCG@20 on all datasets.
The upper block lists centralized models, the middle block contains standard FedSeq baselines, and the lower blocks show ConFedSRS and our LUMOS framework.
The best centralized result in each column is \underline{underlined}, and the best federated result is in \textbf{bold}.

\begin{table*}[t]
\centering
\footnotesize
\caption{Top-$K$ recommendation performance on three datasets.
The best centralized result in each column is \underline{underlined}, and the best federated result is in \textbf{bold}.}
\label{tab:overall-performance}
\vspace{-.5em}
\resizebox{0.8\columnwidth}{!}{%
\begin{tabular}{lcccccc}
\hline
\multirow{2}{*}{Method} &
\multicolumn{2}{c}{Cell Phone} &
\multicolumn{2}{c}{Baby} &
\multicolumn{2}{c}{MIND} \\
 & HR@20 & NDCG@20 & HR@20 & NDCG@20 & HR@20 & NDCG@20 \\
\hline
GRU4Rec            & 0.0700 & 0.0290 & 0.0370 & 0.0140 & 0.1310 & 0.0580 \\
SASRec             & \underline{0.0840} & \underline{0.0370} &
                      \underline{0.0390} & \underline{0.0150} &
                      \underline{0.1510} & \underline{0.0690} \\
\hline
FedSeq(GRU4Rec)    & 0.0500 & 0.0180 & 0.0230 & 0.0088 & 0.0700 & 0.0240 \\
FedSeq(SASRec)     & 0.0510 & 0.0185 & 0.0290 & 0.0102 & 0.0650 & 0.0228 \\
\hline
ConFedSRS(GRU4Rec) & 0.0860 & 0.0385 & 0.0340 & 0.0142 & 0.1300 & 0.0610 \\
ConFedSRS(SASRec)  & 0.0950 & 0.0425 & 0.0420 & 0.0173 & 0.1370 & 0.0620 \\
\hline
\textbf{LUMOS}     & \textbf{0.1570} & \textbf{0.0670} &
                      \textbf{0.0680} & \textbf{0.0250} &
                      \textbf{0.1660} & \textbf{0.0730} \\
\hline
\end{tabular}
}
\vspace{-.5em}
\end{table*}

Moving to the federated setting, both FedSeq variants suffer clear drops in HR@20 and NDCG@20 due to data partitioning and the absence of auxiliary self-supervision.
Adding a simple sequence-level contrastive objective (ConFedSRS) consistently improves over FedSeq, indicating that even handcrafted augmentations help stabilize client representations. LUMOS achieves the best federated performance on all three datasets and metrics, with substantial margins over ConFedSRS.
Compared with the strongest baseline ConFedSRS(SASRec), LUMOS improves HR@20 by $6.2$\,–\,$7.6$ absolute points and NDCG@20 by around $1.0$ point, thanks to the richer tri-view supervision provided by LLM-generated future, paraphrased, and counterfactual behaviors.
Notably, LUMOS also surpasses centralized SASRec on all datasets, suggesting that semantically grounded LLM views can compensate for the lack of direct access to global interactions in federated training.

\begin{table*}[!h]
\centering
\footnotesize
\caption{Ablation study of LUMOS on three datasets. The best results are in bold.}
\label{tab:ablation}
\vspace{-.5em}
\resizebox{0.8\columnwidth}{!}{%
\begin{tabular}{lcccccc}
\hline\noalign{\smallskip}
\multirow{2}{*}{Variant} &
\multicolumn{2}{c}{Cell Phone} &
\multicolumn{2}{c}{Baby} &
\multicolumn{2}{c}{MIND} \\
\cline{2-7}
 & HR@20 & NDCG@20 & HR@20 & NDCG@20 & HR@20 & NDCG@20 \\
\noalign{\smallskip}\hline\noalign{\smallskip}
\textbf{LUMOS (full)} & \textbf{0.1550} & \textbf{0.0660} &
                         \textbf{0.0650} & \textbf{0.0240} &
                         \textbf{0.1620} & \textbf{0.0725} \\
w/o predicted future view      & 0.1410 & 0.0590 & 0.0590 & 0.0215 & 0.1490 & 0.0670 \\
w/o paraphrased intent view          & 0.1360 & 0.0565 & 0.0570 & 0.0205 & 0.1450 & 0.0655 \\
w/o counterfactual view          & 0.1320 & 0.0550 & 0.0555 & 0.0200 & 0.1425 & 0.0648 \\
\noalign{\smallskip}\hline
\end{tabular}
}
\end{table*}

\subsection{Ablation Study (RQ2)}

\autoref{tab:ablation} shows that removing any component of LUMOS leads to a 
noticeable drop in performance. Eliminating the predictive future view causes the 
largest decrease, reducing HR@20 by roughly 0.01--0.015 across datasets. Removing 
the paraphrased or counterfactual views produces similar declines (about 0.008--0.012), 
showing that both provide complementary semantic signals. Dropping the temporal 
regularizer further lowers HR@20 by about 0.005--0.008, confirming its role in 
stabilizing user representations. Overall, the full tri-view design and temporal 
consistency are necessary to achieve the best accuracy.

\subsection{Hyperparameter Analysis (RQ3)}

\begin{figure}[t]
    \vspace{-.5em}
\centering
\includegraphics[width=0.5\linewidth]{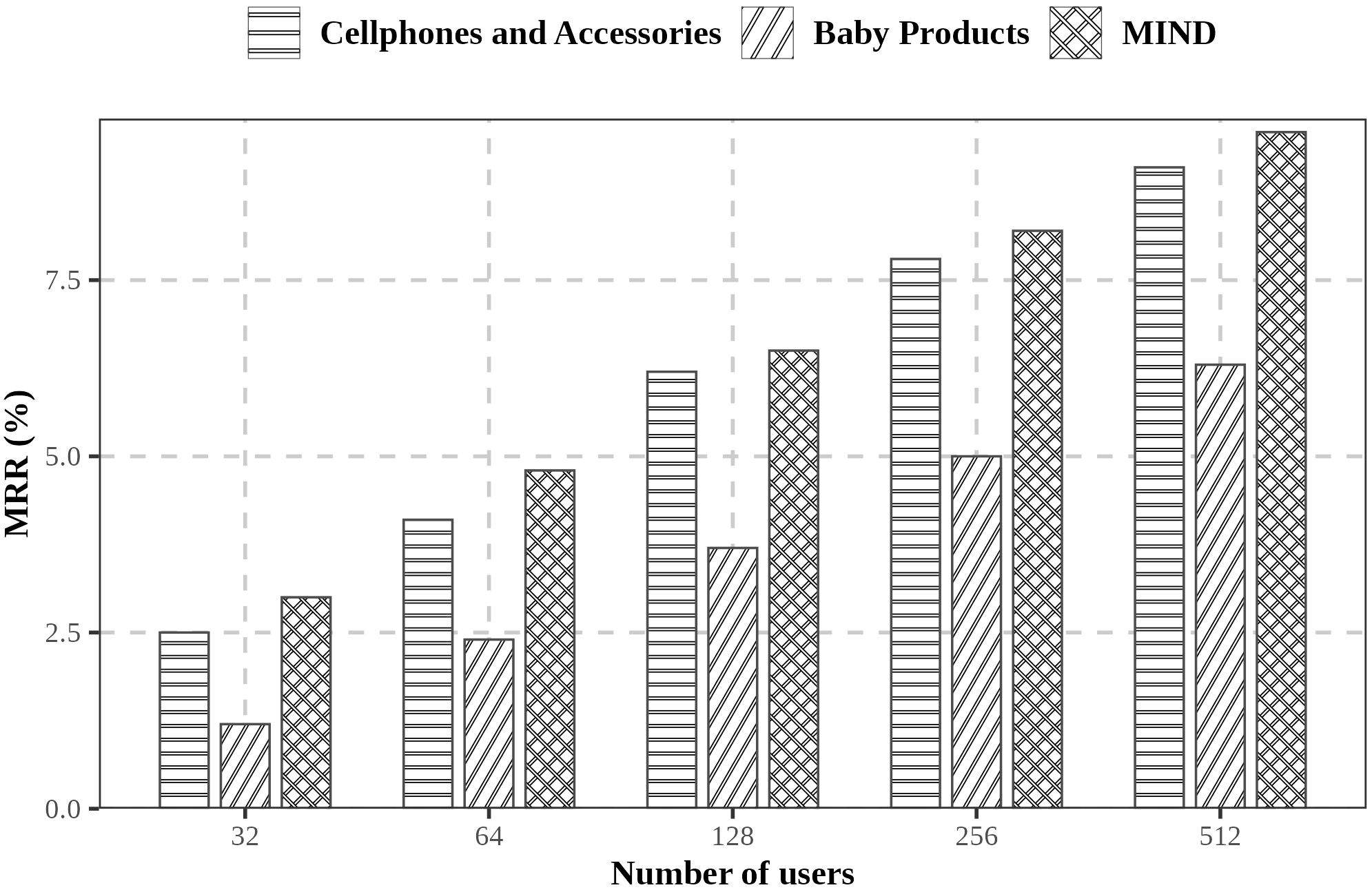}
\vspace{-.5em}
\caption{Impact of the number of participating users on model performance.}
\label{fig:individual_fix}
\vspace{-.5em}
\end{figure}

\sstitle{Effects of number of users every epoch}
As shown in \autoref{fig:individual_fix}, LUMOS improves consistently with larger 
client participation. MRR (Mean Reciprocal Rank), which measures how highly the correct next item is 
ranked for each user, is low at 32 users (about 2--3\%) but rises steadily through 
64 and 128 users, and reaches over 8\% on CellPhone and nearly 9\% on MIND at 
512 users. Federated training also increasingly surpasses individual learning, 
with a gap of more than 3 MRR points at the largest scale. This confirms that 
wider participation provides richer behavioral signals for stronger sequential 
modeling.

\begin{figure}[!h]
    \vspace{-.5em}
\centering
\includegraphics[width=.7\linewidth]{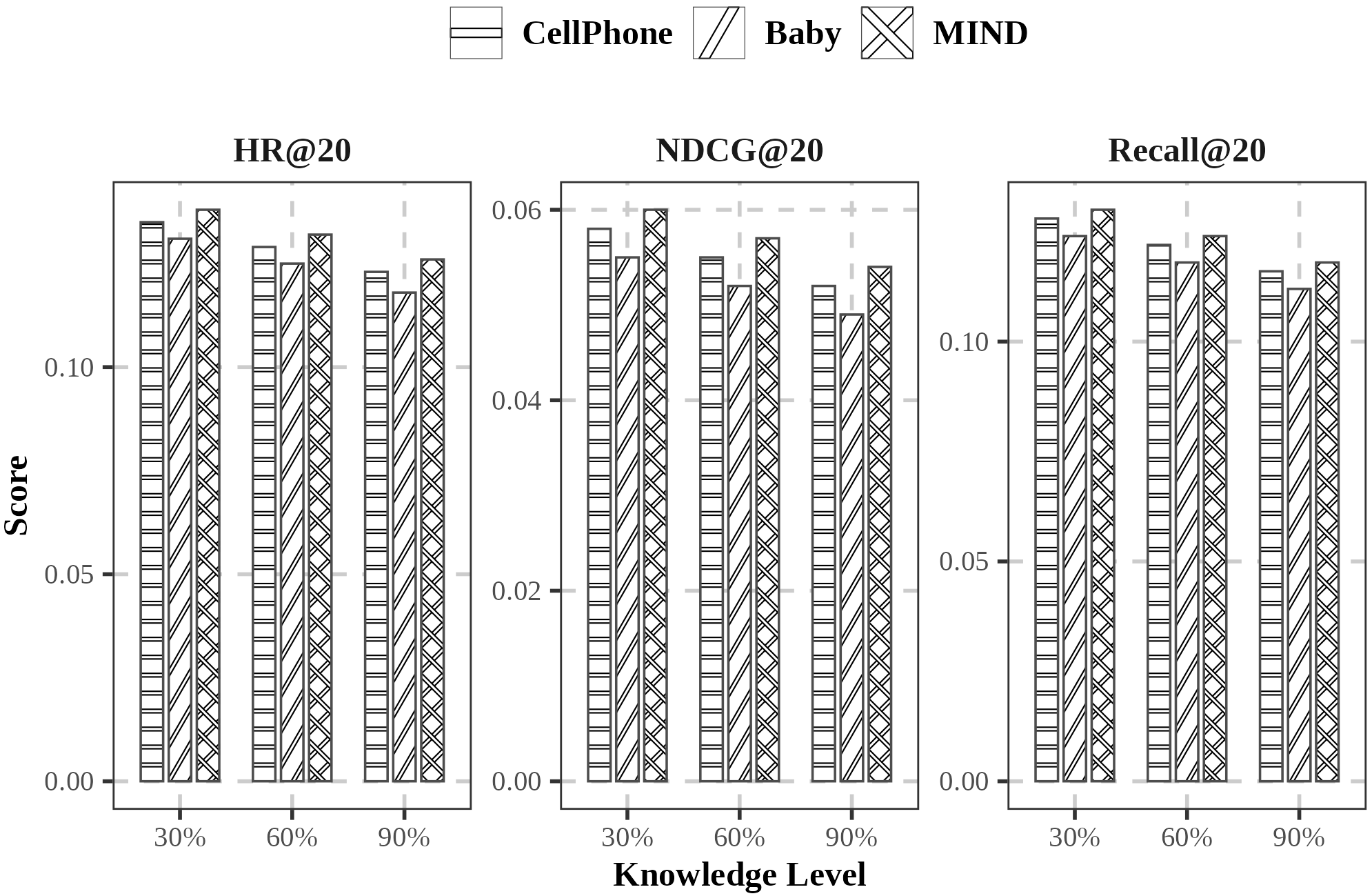}
\vspace{-.5em}
\caption{Effects of partial knowledge on model utility.}
\label{fig:partial_knowledge}
\vspace{-.5em}
\end{figure}

\sstitle{Effects of noise and partial knowledge}
As shown in Figure~\ref{fig:partial_knowledge}, LUMOS remains stable even as 
adversarial knowledge increases from 30\% to 90\%. Across all datasets, HR@20 
drops by only about 0.01--0.015, while NDCG@20 and Recall@20 decrease by less 
than 0.01. These small changes indicate that the semantic structure introduced by 
LLM-based tri-view augmentation effectively limits the impact of informed 
adversarial noise~\cite{pham2024dual}.

\section{Conclusion}
\label{sec:conclusion}

We revisited federated sequential recommendation through the lens of representation learning to address data sparsity and limited augmentations. We proposed LUMOS, which leverages on-device LLMs as \emph{semantic behavioral generators} to enhance local learning.
LUMOS utilizes a tri-view contrastive objective with locally generated predictive, paraphrase, and counterfactual views. This enables a multi-positive InfoNCE loss that aligns semantic behaviors while preserving privacy, avoiding any data leakage to the server.
Future work could explore personalized prompting or adapter tuning~\cite{pham2025multilingual,duong2023deep,sakong2024higher,nguyen2024portable}, and the inclusion of richer signals~\cite{sakong2025handling,nguyen2024multitask,nguyen2022model,nguyen2023validating,nguyen2023detecting}. Furthermore, integrating communication- and energy-aware cost models would facilitate deployment on resource-constrained devices, advancing privacy-preserving LLM-driven federated recommendation~\cite{huynh2025certified,nguyen2025privacy,nguyen2024handling,huynh2024fast,nguyen2023isomorphic}.


\vspace{-2em}

\begin{thebibliography}{10}
\providecommand{\url}[1]{\texttt{#1}}
\providecommand{\urlprefix}{URL }
\providecommand{\doi}[1]{https://doi.org/#1}

\bibitem{rw_24}
Ammad-Ud-Din, M., Ivannikova, E., Khan, S.A., Oyomno, W., Fu, Q., Tan, K.E.,
  Flanagan, A.: Federated collaborative filtering for privacy-preserving
  personalized recommendation system. arXiv preprint arXiv:1901.09888  (2019)

\bibitem{rw_112}
Devlin, J., Chang, M.W., Lee, K., Toutanova, K.: Bert: Pre-training of deep
  bidirectional transformers for language understanding. In: NAACL-HLT. pp.
  4171--4186 (2019)

\bibitem{duong2022deep}
Duong, C.T., Nguyen, T.T., Hoang, T.D., Yin, H., Weidlich, M., Nguyen, Q.V.H.:
  Deep mincut: Learning node embeddings from detecting communities. Pattern
  Recognition p. 109126 (2022)

\bibitem{duong2023deep}
Duong, C.T., Nguyen, T.T., Hoang, T.D., Yin, H., Weidlich, M., Nguyen, Q.V.H.:
  Deep mincut: Learning node embeddings by detecting communities. Pattern
  Recognition  \textbf{134},  109126 (2023)

\bibitem{duong2022efficient}
Duong, C.T., Nguyen, T.T., Yin, H., Weidlich, M., Mai, T.S., Aberer, K.,
  Nguyen, Q.V.H.: Efficient and effective multi-modal queries through
  heterogeneous network embedding. IEEE Transactions on Knowledge and Data
  Engineering  \textbf{34}(11),  5307--5320 (2022)

\bibitem{rw_15}
Fang, H., Zhang, D., Shu, Y., Guo, G.: Deep learning for sequential
  recommendation: Algorithms, influential factors, and evaluations. TOIS
  \textbf{39}(1),  1--42 (2020)

\bibitem{rw_11}
He, X., Deng, K., Wang, X., Li, Y., Zhang, Y., Wang, M.: Lightgcn: Simplifying
  and powering graph convolution network for recommendation. In: SIGIR. pp.
  639--648 (2020)

\bibitem{rw_12}
He, X., Liao, L., Zhang, H., Nie, L., Hu, X., Chua, T.S.: Neural collaborative
  filtering. In: WWW. pp. 173--182 (2017)

\bibitem{hung2017answer}
Hung, N.Q.V., Thang, D.C., Tam, N.T., Weidlich, M., Aberer, K., Yin, H., Zhou,
  X.: Answer validation for generic crowdsourcing tasks with minimal efforts.
  The VLDB Journal  \textbf{26},  855--880 (2017)

\bibitem{hung2019handling}
Hung, N.Q.V., Weidlich, M., Tam, N.T., Mikl{\'o}s, Z., Aberer, K., Gal, A.,
  Stantic, B.: Handling probabilistic integrity constraints in pay-as-you-go
  reconciliation of data models. Information Systems  \textbf{83},  166--180
  (2019)

\bibitem{huynh2021network}
Huynh, T.T., Duong, C.T., Nguyen, T.T., Van, V.T., Sattar, A., Yin, H., Nguyen,
  Q.V.H.: Network alignment with holistic embeddings. TKDE  \textbf{35}(2),
  1881--1894 (2021)

\bibitem{huynh2023efficient}
Huynh, T.T., Nguyen, M.H., Nguyen, T.T., Nguyen, P.L., Weidlich, M., Nguyen,
  Q.V.H., Aberer, K.: Efficient integration of multi-order dynamics and
  internal dynamics in stock movement prediction. In: Proceedings of the
  Sixteenth ACM International Conference on Web Search and Data Mining. pp.
  850--858 (2023)

\bibitem{huynh2024fast}
Huynh, T.T., Nguyen, T.B., Nguyen, P.L., Nguyen, T.T., Weidlich, M., Nguyen,
  Q.V.H., Aberer, K.: Fast-fedul: A training-free federated unlearning with
  provable skew resilience. In: ECML-PKDD. pp. 55--72 (2024)

\bibitem{huynh2025certified}
Huynh, T.T., Nguyen, T.B., Nguyen, T.T., Nguyen, P.L., Yin, H., Nguyen, Q.V.H.,
  Nguyen, T.T.: Certified unlearning for federated recommendation. TOIS
  \textbf{43}(2),  1--29 (2025)

\bibitem{rw_16}
Jannach, D., Ludewig, M.: When recurrent neural networks meet the neighborhood
  for session-based recommendation. In: RecSys. pp. 306--310 (2017)

\bibitem{rw_49}
Jing, M., Zhu, Y., Zang, T., Wang, K.: Contrastive self-supervised learning in
  recommender systems: A survey. TOIS  \textbf{42}(2),  1--39 (2023)

\bibitem{rw_17}
Kang, W.C., McAuley, J.: Self-attentive sequential recommendation. In: ICDM.
  pp. 197--206 (2018)

\bibitem{krichene2020sampled}
Krichene, W., Rendle, S.: On sampled metrics for item recommendation. In: KDD.
  pp. 1748--1757 (2020)

\bibitem{fmss}
Lin, Z., Pan, W., Yang, Q., Ming, Z.: A generic federated recommendation
  framework via fake marks and secret sharing. TOIS  \textbf{41}(2),  1--37
  (2022)

\bibitem{rw_21}
Liu, Y., Tan, B., Zheng, V.W., Chen, K., Yang, Q.: Federated recommendation
  systems. In: Federated Learning: Privacy and Incentive, pp. 225--239 (2020)

\bibitem{long2024physical}
Long, J., Chen, T., Ye, G., Zheng, K., Nguyen, Q.V.H., Yin, H.: Physical
  trajectory inference attack and defense in decentralized poi recommendation.
  In: WWW. pp. 3379--3387 (2024)

\bibitem{rw_23}
Long, J., Ye, G., Chen, T., et~al.: Diffusion-based cloud-edge-device
  collaborative learning for next poi recommendations. In: KDD. pp. 2026--2036
  (2024)

\bibitem{rw_54}
Luo, L., Liu, B.: Dual-contrastive for federated social recommendation. In:
  IJCNN. pp.~1--8 (2022)

\bibitem{rw_53}
Luo, S., Xiao, Y., Zhang, X., Liu, Y., Ding, W., Song, L.: Perfedrec++:
  Enhancing personalized federated recommendation with self-supervised
  pre-training. TIST  \textbf{15}(5),  1--24 (2024)

\bibitem{llmrec}
Lyu, H., Jiang, S., Zeng, H., Xia, Y., Wang, Q., Zhang, S., Chen, R., Leung,
  C., Tang, J., Luo, J.: {LLM}-rec: Personalized recommendation via prompting
  large language models. In: NAACL. pp. 583--612 (2024)

\bibitem{rw_110}
Min, B., Ross, H., Sulem, E., Veyseh, A.P.B., Nguyen, T.H., Sainz, O., Agirre,
  E., Heintz, I., Roth, D.: Recent advances in natural language processing via
  large pre-trained language models: A survey. CSUR  \textbf{56}(2),  1--40
  (2023)

\bibitem{rw_52}
Minto, L., Haller, M., Livshits, B., et~al.: Stronger privacy for federated
  collaborative filtering with implicit feedback. In: RecSys. pp. 342--350
  (2021)

\bibitem{nguyen2024multitask}
Nguyen, D.D.A., Nguyen, M.H., Nguyen, P.L., Jo, J., Yin, H., Nguyen, T.T.:
  Multi-task learning of heterogeneous hypergraph representations in {LBSNs}.
  In: ADMA. pp. 161--177 (2024)

\bibitem{nguyen2025ondevice}
Nguyen, M.H., Huynh, T.T., Nguyen, T.T., Nguyen, T., Nguyen, P.L., Pham, H.T.,
  Jo, J.: On-device diagnostic recommendation with heterogeneous federated
  blocknets. SCIS  \textbf{68}(4),  140102 (2025)

\bibitem{nguyen2015tag}
Nguyen, Q.V.H., Do, S.T., Nguyen, T.T., Aberer, K.: Tag-based paper retrieval:
  minimizing user effort with diversity awareness. In: International Conference
  on Database Systems for Advanced Applications. pp. 510--528 (2015)

\bibitem{rw_13}
Nguyen, Q.V.H., Duong, C.T., Nguyen, T.T., Weidlich, M., Aberer, K., Yin, H.,
  Zhou, X.: Argument discovery via crowdsourcing. VLDBJ  \textbf{26},  511--535
  (2017)

\bibitem{nguyen2017argument}
Nguyen, Q.V.H., Duong, C.T., Nguyen, T.T., Weidlich, M., Aberer, K., Yin, H.,
  Zhou, X.: Argument discovery via crowdsourcing. The VLDB Journal
  \textbf{26}(4),  511--535 (2017)

\bibitem{nguyen2015smart}
Nguyen, Q.V.H., Nguyen, T.T., Chau, V.T., Wijaya, T.K., Mikl{\'o}s, Z., Aberer,
  K., Gal, A., Weidlich, M.: Smart: A tool for analyzing and reconciling schema
  matching networks. In: ICDE. pp. 1488--1491 (2015)

\bibitem{nguyen2014reconciling}
Nguyen, Q.V.H., Nguyen~Thanh, T., Mikl{\'o}s, Z., Aberer, K.: Reconciling
  schema matching networks through crowdsourcing. EAI Endorsed Transactions on
  Collaborative Computing  \textbf{1}(2), ~e2 (2014)

\bibitem{nguyen2018if}
Nguyen, Q.V.H., Zheng, K., Weidlich, M., Zheng, B., Yin, H., Nguyen, T.T.,
  Stantic, B.: What-if analysis with conflicting goals: Recommending data
  ranges for exploration. In: 2018 IEEE 34th International Conference on Data
  Engineering (ICDE). pp. 89--100. IEEE (2018)

\bibitem{rw_31}
Nguyen, T.T., Nguyen, Q.V.H., Nguyen, T.T., et~al.: Manipulating recommender
  systems: a survey of poisoning attacks and countermeasures. CSUR
  \textbf{57},  1--39 (2025)

\bibitem{nguyen2023validating}
Nguyen, T.T., Huynh, T.T., Pham, M.T., Hoang, T.D., Nguyen, T.T., Nguyen,
  Q.V.H.: Validating functional redundancy with mixed generative adversarial
  networks. KBS  \textbf{264},  110342 (2023)

\bibitem{nguyen2025privacy}
Nguyen, T.T., Huynh, T.T., Ren, Z., Nguyen, T.T., Nguyen, P.L., Yin, H.,
  Nguyen, Q.V.H.: Privacy-preserving explainable ai: a survey. SCIS
  \textbf{68}(1),  111101 (2025)

\bibitem{nguyen2023detecting}
Nguyen, T.T., Huynh, T.T., Yin, H., Weidlich, M., Nguyen, T.T., Mai, T.S.,
  Nguyen, Q.V.H.: Detecting rumours with latency guarantees using massive
  streaming data. VLDBJ  \textbf{32}(2),  369--387 (2023)

\bibitem{nguyen2024handling}
Nguyen, T.T., Nguyen, T.T., Weidlich, M., Jo, J., Nguyen, Q.V.H., Yin, H.,
  Liew, A.W.C.: Handling low homophily in recommender systems with partitioned
  graph transformer. TKDE  (2024)

\bibitem{nguyen2022model}
Nguyen, T.T., Phan, T.C., Nguyen, M.H., Weidlich, M., Yin, H., Jo, J., Nguyen,
  Q.V.H.: Model-agnostic and diverse explanations for streaming rumour graphs.
  KBS  \textbf{253},  109438 (2022)

\bibitem{nguyen2023example}
Nguyen, T.T., Phan, T.C., Pham, H.T., Nguyen, T.T., Jo, J., Nguyen, Q.V.H.:
  Example-based explanations for streaming fraud detection on graphs.
  Information Sciences  \textbf{621},  319--340 (2023)

\bibitem{nguyen2024portable}
Nguyen, T.T., Ren, Z., Nguyen, T.T., Jo, J., Nguyen, Q.V.H., Yin, H.: Portable
  graph-based rumour detection against multi-modal heterophily. KBS
  \textbf{284},  111310 (2024)

\bibitem{nguyen2026review}
Nguyen, T.T., Ren, Z., Pham, T., Nguyen, P.L., Nguyen, Q.V.H., Yin, H.: A
  review of instruction-guided image editing. EAAI  (2026)

\bibitem{nguyen2020factcatch}
Nguyen, T.T., Weidlich, M., Yin, H., Zheng, B., Nguyen, Q.H., Nguyen, Q.V.H.:
  Factcatch: Incremental pay-as-you-go fact checking with minimal user effort.
  In: Proceedings of the 43rd International ACM SIGIR Conference on Research
  and Development in Information Retrieval. pp. 2165--2168 (2020)

\bibitem{nguyen2023isomorphic}
Nguyen, T.T., Nguyen, T.T., Nguyen, T.H., Yin, H., Nguyen, T.T., Jo, J.,
  Nguyen, Q.V.H.: Isomorphic graph embedding for progressive maximal frequent
  subgraph mining. TIST  \textbf{15}(1),  1--26 (2023)

\bibitem{nguyen2023poisoning}
Nguyen~Thanh, T., Quach, N.D.K., Nguyen, T.T., Huynh, T.T., Vu, V.H., Nguyen,
  P.L., Jo, J., Nguyen, Q.V.H.: Poisoning gnn-based recommender systems with
  generative surrogate-based attacks. ACM Transactions on Information Systems
  \textbf{41}(3),  1--24 (2023)

\bibitem{dataset_amazon}
Ni, J., Li, J., McAuley, J.: Justifying recommendations using distantly-labeled
  reviews and fine-grained aspects. In: EMNLP-IJCNLP. pp. 188--197 (2019)

\bibitem{pham2025multilingual}
Pham, K.T., Nguyen, T.H., Jo, J., Nguyen, Q.V.H., Nguyen, T.T.: Multilingual
  text-to-sql: Benchmarking the limits of language models with collaborative
  language agents. In: ADC. pp. 108--123 (2025)

\bibitem{pham2024dual}
Pham, M.T., Huynh, T.T., Nguyen, T.T., Nguyen, T.T., Jo, J., Yin, H., Nguyen,
  Q.V.H.: A dual benchmarking study of facial forgery and facial forensics.
  CAAI-TIT  \textbf{9}(6),  1377--1397 (2024)

\bibitem{rw_25}
Qu, L., Tang, N., Zheng, R., Nguyen, Q.V.H., Huang, Z., Shi, Y., Yin, H.:
  Semi-decentralized federated ego graph learning for recommendation. In: WWW.
  pp. 339--348 (2023)

\bibitem{qu2024towards}
Qu, L., Yuan, W., Zheng, R., Cui, L., Shi, Y., Yin, H.: Towards personalized
  privacy: User-governed data contribution for federated recommendation. In:
  WWW. pp. 3910--3918 (2024)

\bibitem{ren2022prototype}
Ren, Z., Nguyen, T.T., Nejdl, W.: {Prototype learning for interpretable
  respiratory sound analysis}. In: Proc.\ ICASSP. pp. 9087--9091 (2022)

\bibitem{rw_14}
Rendle, S., Freudenthaler, C., Schmidt-Thieme, L.: Factorizing personalized
  markov chains for next-basket recommendation. In: WWW. pp. 811--820 (2010)

\bibitem{a-hum}
Rong, D., He, Q., Chen, J.: Poisoning deep learning based recommender model in
  federated learning scenarios. In: IJCAI. pp. 2204--2210 (2022)

\bibitem{sakong2025handling}
Sakong, D., Nguyen, T.T.: Handling heterophily in recommender systems with
  wavelet hypergraph diffusion. arXiv preprint arXiv:2501.14399  (2025)

\bibitem{sakong2024higher}
Sakong, D., Vu, V.H., Huynh, T.T., Le~Nguyen, P., Yin, H., Nguyen, Q.V.H.,
  Nguyen, T.T.: Higher-order knowledge-enhanced recommendation with
  heterogeneous hypergraph multi-attention. Information Sciences  \textbf{680},
   121165 (2024)

\bibitem{fi17060252}
Shehmir, S., Kashef, R.: {LLM4Rec}: A comprehensive survey on the integration
  of large language models in recommender systems---approaches, applications
  and challenges. Future Internet  \textbf{17}(6) (2025)

\bibitem{rw_18}
Sun, F., Liu, J., Wu, J., Pei, C., Lin, X., Ou, W., Jiang, P.: Bert4rec:
  Sequential recommendation with bidirectional encoder representations from
  transformer. In: CIKM. pp. 1441--1450 (2019)

\bibitem{tang2025}
Tang, Z., Huan, Z., Li, Z., Zhang, X., Hu, J., Fu, C., Zhou, J., Zou, L., Li,
  C.: One model for all: Large language models are domain-agnostic
  recommendation systems. TOIS  \textbf{43}(5),  1--27 (2025)

\bibitem{thang2022nature}
Thang, D.C., Dat, H.T., Tam, N.T., Jo, J., Hung, N.Q.V., Aberer, K.: Nature vs.
  nurture: Feature vs. structure for graph neural networks. PRL  \textbf{159},
  46--53 (2022)

\bibitem{thang2015evaluation}
Thang, D.C., Tam, N.T., Hung, N.Q.V., Aberer, K.: An evaluation of
  diversification techniques. In: International Conference on Database and
  Expert Systems Applications. pp. 215--231 (2015)

\bibitem{toan2018diversifying}
Toan, N.T., Cong, P.T., Tam, N.T., Hung, N.Q.V., Stantic, B.: Diversifying
  group recommendation. IEEE Access  \textbf{6},  17776--17786 (2018)

\bibitem{trung2022learning}
Trung, H.T., Van~Vinh, T., Tam, N.T., Jo, J., Yin, H., Hung, N.Q.V.: Learning
  holistic interactions in lbsns with high-order, dynamic, and multi-role
  contexts. IEEE Transactions on Knowledge and Data Engineering
  \textbf{35}(5),  5002--5016 (2022)

\bibitem{rw_46}
Wang, H., Xu, Y., Yang, C., Shi, C., Li, X., Guo, N., Liu, Z.:
  Knowledge-adaptive contrastive learning for recommendation. In: WSDM. pp.
  535--543 (2023)

\bibitem{rw_32}
Wang, Z., Yu, J., Gao, M., et~al.: Poisoning attacks and defenses in
  recommender systems: a survey. arXiv preprint arXiv:2406.01022  (2024)

\bibitem{rw_51}
Wu, C., Wu, F., Qi, T., et~al.: Fedcl: Federated contrastive learning for
  privacy-preserving recommendation. arXiv preprint arXiv:2204.09850  (2022)

\bibitem{dataset_mind}
Wu, F., Qiao, Y., Chen, J.H., Wu, C., Qi, T., Lian, J., Liu, D., Xie, X., Gao,
  J., Wu, W., et~al.: Mind: A large-scale dataset for news recommendation. In:
  ACL. pp. 3597--3606 (2020)

\bibitem{rw_412}
Wu, J., Wang, X., Feng, F., et~al.: Self-supervised graph learning for
  recommendation. In: SIGIR. pp. 726--735 (2021)

\bibitem{wu2024llm4rec_survey}
Wu, L., Zheng, Z., Qiu, Z., Wang, H., Gu, H., Shen, T., Qin, C., Zhu, C., Zhu,
  H., Liu, Q., Xiong, H., Chen, E.: A survey on large language models for
  recommendation. WWW  \textbf{27}(5), ~60 (2024)

\bibitem{rw_45}
Wu, Y., Xie, R., Zhu, Y., Ao, X., Chen, X., Zhang, X., Zhuang, F., Lin, L., He,
  Q.: Multi-view multi-behavior contrastive learning in recommendation. In:
  DASFAA. pp. 166--182 (2022)

\bibitem{rw_27}
Wu, Z., Pan, S., Chen, F., Long, G., Zhang, C., Yu, P.S.: A comprehensive
  survey on graph neural networks. TNNLS  \textbf{32}(1),  4--24 (2020)

\bibitem{rw_410}
Xia, X., Yin, H., Yu, J., et~al.: Self-supervised hypergraph convolutional
  networks for session-based recommendation. In: AAAI. pp. 4503--4511 (2021)

\bibitem{rw_44}
Xie, X., Sun, F., Liu, Z., Wu, S., Gao, J., Zhang, J., Ding, B., Cui, B.:
  Contrastive learning for sequential recommendation. In: ICDE. pp. 1259--1273
  (2022)

\bibitem{yao2024}
Yao, Y., Zhang, J., Wu, J., Huang, C., Xia, Y., Yu, T., Zhang, R., Kim, S.,
  Rossi, R., Li, A., Yao, L., McAuley, J., Chen, Y., Joe-Wong, C.: Federated
  large language models: Current progress and future directions (2024)

\bibitem{ye2024}
Ye, R., Ge, R., Zhu, X., Chai, J., Du, Y., Liu, Y., Wang, Y., Chen, S.:
  {FedLLM-Bench}: Realistic benchmarks for federated learning of large language
  models (2024)

\bibitem{rw_22}
Yin, H., Qu, L., Chen, T., Yuan, W., Zheng, R., Long, J., Xia, X., Shi, Y.,
  Zhang, C.: On-device recommender systems: A comprehensive survey. arXiv
  preprint arXiv:2401.11441  (2024)

\bibitem{rw_41}
Yu, J., Xia, X., Chen, T., Cui, L., Nguyen, Q.V.H., Yin, H.: Xsimgcl: Towards
  extremely simple graph contrastive learning for recommendation. TKDE  (2023)

\bibitem{rw_411}
Yu, J., Yin, H., Li, J., et~al.: Self-supervised multi-channel hypergraph
  convolutional network for social recommendation. In: WWW. pp. 413--424 (2021)

\bibitem{rw_42}
Yu, J., Yin, H., Xia, X., Chen, T., Li, J., Huang, Z.: Self-supervised learning
  for recommender systems: A survey. TKDE  (2023)

\bibitem{rw_415}
Yu, J., Yin, H., Xia, X., et~al.: Are graph augmentations necessary? simple
  graph contrastive learning for recommendation. In: SIGIR. pp. 1294--1303
  (2022)

\bibitem{rw_37}
Yu, Y., Liu, Q., Wu, L., et~al.: Untargeted attack against federated
  recommendation systems via poisonous item embeddings and the defense. In:
  AAAI. pp. 4854--4863 (2023)

\bibitem{psmu}
Yuan, W., Nguyen, Q.V.H., He, T., et~al.: Manipulating federated recommender
  systems: poisoning with synthetic users and its countermeasures. In: SIGIR.
  pp. 1690--1699 (2023)

\bibitem{rw_26}
Yuan, W., Qu, L., Cui, L., Tong, Y., Zhou, X., Yin, H.: Hetefedrec: Federated
  recommender systems with model heterogeneity. arXiv preprint arXiv:2307.12810
   (2023)

\bibitem{rw_210}
Yuan, W., Yang, C., Nguyen, Q.V.H., Cui, L., He, T., Yin, H.: Interaction-level
  membership inference attack against federated recommender systems. In: WWW.
  pp. 1053--1062 (2023)

\bibitem{ptf-fsr}
Yuan, W., Yang, C., Qu, L., Hung~Nguyen, Q.V., Ye, G., Yin, H.: Ptf-fsr: A
  parameter transmission-free federated sequential recommender system. TOIS
  \textbf{43}(2) (2025)

\bibitem{rw_211}
Yuan, W., Yang, C., Qu, L., et~al.: Hide your model: a parameter
  transmission-free federated recommender system. arXiv preprint
  arXiv:2311.14968  (2023)

\bibitem{rw_111}
Yuan, Z., Yuan, F., Song, Y., Li, Y., Fu, J., Yang, F., Pan, Y., Ni, Y.: Where
  to go next for recommender systems? id-vs. modality-based recommender models
  revisited. In: SIGIR. pp. 2639--2649 (2023)

\bibitem{rw_212}
Zhang, S., Yuan, W., Yin, H.: Comprehensive privacy analysis on federated
  recommender system against attribute inference attacks. TKDE  (2023)

\bibitem{zhao2021eires}
Zhao, B., van~der Aa, H., Nguyen, T.T., Nguyen, Q.V.H., Weidlich, M.: Eires:
  Efficient integration of remote data in event stream processing. In:
  Proceedings of the 2021 International Conference on Management of Data. pp.
  2128--2141 (2021)

\bibitem{rw_113}
Zhao, Z., Fan, W., Li, J., Liu, Y., Mei, X., Wang, Y., Wen, Z., Wang, F., Zhao,
  X., Tang, J., et~al.: Recommender systems in the era of large language models
  (llms). TKDE  (2024)

\end{thebibliography}

\end{document}